\documentclass[
]{ceurart}

\usepackage{listings}
\usepackage{graphicx}
\usepackage{subcaption}
\usepackage{caption}

\begin{document}

% \copyrightyear{2025}
% \copyrightclause{Copyright for this paper by its authors.
%   Use permitted under Creative Commons License Attribution 4.0
%   International (CC BY 4.0).}
% \conference{ISWC'25: International Semantic Web Conference, November 2--6,  2025, Nara, Japan}

\copyrightyear{2025}
\copyrightclause{Copyright for this paper by its authors. Use permitted under Creative Commons License Attribution 4.0 International (CC BY 4.0).}
\conference{ISWC 2025 Companion Volume, November 2--6, 2025, Nara, Japan}

\title{Bio-KGvec2go: Serving up-to-date Dynamic Biomedical Knowledge Graph Embeddings}

\tnotemark[1]
\tnotetext[1]{You can use this document as the template for preparing your
  publication. We recommend using the latest version of the ceurart style.}

\author[1]{Hamid Ahmad}
\address[1]{University of Mannheim, Mannheim, Germany}

\author[2]{Heiko Paulheim}

\author[2]{Rita T. Sousa}[%
orcid=0000-0002-7241-8970,
email=rita.sousa@uni-mannheim.de,
]
\cormark[1]
\address[2]{Data and Web Science Group, University of Mannheim, Mannheim, Germany}

% \author[4]{Manfred Jeusfeld}[%
% orcid=0000-0002-9421-8566,
% email=Manfred.Jeusfeld@acm.org,
% url=http://conceptbase.sourceforge.net/mjf/,
% \address[4]{University of Skövde, Högskolevägen 1, 541 28 Skövde, Sweden}

% %% Footnotes
% \cortext[1]{Corresponding author.}
% \fntext[1]{These authors contributed equally.}

\begin{abstract}
Knowledge graphs and ontologies represent entities and their relationships in a structured way, having gained significance in the development of modern AI applications. Integrating these semantic resources with machine learning models often relies on knowledge graph embedding models to transform graph data into numerical representations. Therefore, pre-trained models for popular knowledge graphs and ontologies are increasingly valuable, as they spare the need to retrain models for different tasks using the same data, thereby helping to democratize AI development and enabling sustainable computing. 

In this paper, we present Bio-KGvec2go, an extension of the KGvec2go Web API, designed to generate and serve knowledge graph embeddings for widely used biomedical ontologies. Given the dynamic nature of these ontologies, Bio-KGvec2go also supports regular updates aligned with ontology version releases. By offering up-to-date embeddings with minimal computational effort required from users, Bio-KGvec2go facilitates efficient and timely biomedical research.

\end{abstract}


\maketitle

%%%%%%%%%%%%%%%%%%%%%%%% INTRODUCTION %%%%%%%%%%%%%%%%%%%%%%%%
\section{Introduction}

% KGs and ontologies 
Knowledge Graphs (KGs) contain factual knowledge about real-world entities and their relations in a fully machine-readable format~\cite{hogan2021knowledge}. Many modern KGs represent this information according to a formal definition of the domain knowledge given by an ontology. Ontologies are semantic models for a domain in which each entity is precisely defined, and the relationships between entities are parameterized or constrained~\cite{staab2013handbook}. 
In life sciences, the use of ontologies has gained prominence, with increasing importance in biomedical research~\cite{hoehndorf2013evaluation}. Ontologies are applied across various areas of biology and medicine, ranging from gene function~\cite{gene2019gene} to drug characterization~\cite{chebi2007}. Phenotype ontologies are also available for multiple species for the characterization of diseases~\cite{kohler2021human}. Open repositories, such as BioPortal~\cite{bioportal2011}, provide access to hundreds of biomedical ontologies.

% Use of KG embedding methods 
Given the richness of these semantic resources, they have been exploited in a wide variety of machine learning (ML) tasks, including entity classification, link prediction, graph classification, and relation prediction, among others. 
One of the challenges faced by approaches that combine artificial intelligence (AI) with KGs and ontologies is transforming graph data into a suitable representation that can be processed by ML algorithms. A current major trend is the use of KG embedding (KGE) methods, which transform entities and relationships in a KG into a lower-dimensional vector space while attempting to preserve the graph structure and, in some cases, semantic information~\cite{wang2017knowledge}. KGEs are then fed as features for ML algorithms to support several applications, with particular success in the life sciences. From finding new treatments for existing drugs to diagnosing patients and identifying associations between diseases and genes, KGEs have been employed in a wide range of biomedical applications~\cite{mohamed2021biological}.

% Pre-trained
Therefore, pre-trained embeddings for popular biomedical ontologies are increasingly valuable, sparing the need to retrain the models for different tasks using the same data, and allowing greener computing and sustainable AI development.
However, such pre-trained models are not always readily available for the biomedical domain, and when available, they typically reflect a static snapshot of the KG at a specific point in time. 
This poses a challenge, given that knowledge is constantly evolving. Discoveries are published daily, rendering some facts obsolete and revealing new knowledge. As a result, the development of KGs and ontologies is dynamic~\cite{flouris2008ontology}. Relying on embeddings generated on outdated versions risks overlooking critical insights and recent advances, limiting downstream performance.

% Proposal
This paper addresses the challenge of providing up-to-date embeddings for the two most successful biomedical ontologies - Gene Ontology~\cite{gene2019gene} and Human Phenotype Ontology~\cite{kohler2021human}.
We present a framework capable of periodically collecting new KG versions, computing embeddings, and making them publicly available to support downstream research. 
These embeddings are provided via the Bio-KGvec2go platform, \url{www.bio.kgvec2go.org},
%\footnote{\url{http://www.bio.kgvec2go.org}},
which is built upon KGVec2go~\cite{portisch2020kgvec2go}. KGvec2go provides a Web API that enables access to embeddings, computes entity similarity, and identifies related concepts based on input embeddings. While KGvec2go makes available RDF2Vec embeddings~\cite{ristoski2016rdf2vec} for four general-purpose KGs (ALOD, DBpedia, WordNet, and Caligraph), it does not support biomedical KGs or reflect KG evolution. Bio-KGvec2go expands the range of KGs to encompass biomedical ontologies and other KGE models, but also recomputes the embeddings when new ontology versions are released. 

By publicly providing regularly updated and accessible KGEs, we aim to facilitate ongoing research and democratize access to these resources. Even researchers without computational power to train KGE models can conduct analyses and investigations with the latest data representations. Furthermore, it facilitates the study of knowledge evolution, allowing researchers to explore how changes across KG versions impact the resulting embeddings and reflect shifts in domain knowledge over time.

The remainder of this paper is structured as follows: we first describe the two biomedical ontologies explored, followed by an overview of the KGE models employed. Finally, we present the implementation details and functionalities of the Bio-KGvec2go platform.

%%%%%%%%%%%%%%%%%%%%%%%% DATASETS %%%%%%%%%%%%%%%%%%%%%%%%
\section{Biomedical Ontologies}

Currently, Bio-KGvec2go focuses on providing embeddings for two widely used biomedical ontologies. 

\paragraph{Gene Ontology (GO)}
GO~\cite{gene2019gene} defines a hierarchy of more than 40 000 classes that describe protein functions and their relationships.
It can be represented as a graph where nodes are GO classes and edges define relationships between them (e.g., $is\_a$, $part\_of$, $regulates$), with the majority of $is\_a$ relations. 
Functions in GO are described across three domains: the biological processes, the molecular functions, and the cellular components.
GO was initially proposed in 1998 by a consortium of researchers. Since then, it has been constantly reviewed, including the addition or deprecation of terms and reorganization of the relationship structure. 
Most revisions result from advances in biological knowledge or improvements in the precision of experimental technologies. Official GO versions are released monthly\footnote{\url{https://release.geneontology.org/}}.
GO embeddings have been widely used in multiple applications, such as protein function prediction~\cite{zhong2020graph}, protein interactions prediction~\cite{chen2019protein,teremie2022TransformerGO}, and gene-disease associations discovery~\cite{nunes2023multi}. 

\paragraph{Human Phenotype Ontology (HP)}
HP~\cite{kohler2021human} characterizes the phenotypic abnormalities in human hereditary diseases, covering key aspects such as the phenotypic abnormalities themselves, past medical history, mode of inheritance, clinical course, clinical modifiers, and frequency. 
The HP contains more than 18 000 classes represented in a directed acyclic graph, where each node represents a distinct phenotype, and all relationships are of the type $is\_a$, establishing a hierarchy. 
HP was initially developed in 2008 at the Charité University Hospital in Berlin, and it has been continuously updated through a combination of expert curation, integration of new findings from the biomedical literature, and feedback from the global community of clinicians and researchers who use it. While HP does not follow a formal monthly release model like GO, new official versions are made available regularly (approximately every month to two months) through its GitHub repository\footnote{\url{https://github.com/obophenotype/human-phenotype-ontology/releases}}. 
HP embeddings have also been employed in critical biomedical tasks, including patient similarity computation~\cite{shen2019hpo2vec+}, genotype-phenotype association prediction~\cite {patel2022graph}, and gene-disease association prediction~\cite{mukherjee2021identifying}.

%%%%%%%%%%%%%%%%%%%%%%%% KG EMBEDDING MODELS %%%%%%%%%%%%%%%%%%%%%%%%
\section{Knowledge Graph Embedding Models}

KGE methods map each node to a lower-dimensional space where the underlying KG structure and other semantic information are preserved as much as possible. Numerous KGE models have been proposed in the literature, contributing to a substantial body of work as highlighted in different surveys~\cite{cao2024knowledge}. 

The KGE methods can be broadly categorized based on their underlying mechanisms for capturing graph structure and semantic information: translational distance models interpret relations as vector translations, semantic matching models focus on similarity scoring, geometric models exploit spatial structures to encode logical constraints, and random walks-based models use paths through the graph to capture long distance relationships. While translational distance models and semantic matching models focus on exploring the KG triples solely, random walks-based and geometric models also include additional information, namely the hierarchical information.
To encompass a diverse range of knowledge representation approaches, this work employs six representative KGE models spanning the different categories: 

\begin{itemize}
    \item \textbf{TransE}~\cite{bordes2013translating} is the most representative translational distance model, treating relations as vector translations between entities. However, TransE struggles to handle one-to-many and many-to-many relations. Tackling this, \textbf{TransR}~\cite{lin2015learning} introduces a space for each relation.
    
    \item Semantic matching approaches exploit similarity between entities and relations. \textbf{DistMult}~\cite{yang2014embedding} achieves this by employing a bilinear scoring function with diagonal relation matrices. \textbf{HolE}~\cite{nickel2016holographic} uses circular correlation to create compositional representations while remaining scalable.

    \item \textbf{RDF2Vec}~\cite{ristoski2016rdf2vec} is a random walk-based approach built upon two main steps: (i) producing random walks in the graph that are akin to a corpus of sentences; (2) using those sequences as input to a neural language model that learns a latent low-dimensional representation. 

    \item \textbf{BoxE}~\cite{abboud2020boxe} is a geometric approach that represents entities as points, and relations as a set of hyper-rectangles (boxes), capturing logical patterns such as hierarchy, symmetry, or intersection. 
\end{itemize}

% The PyKEEN implementation\footnote{\url{https://pykeen.readthedocs.io/en/stable/}} is used to train TransE, TransR, DistMult, HolE, and BoxE, while pyRDF2Vec\footnote{\url{https://pyrdf2vec.readthedocs.io/en/latest/}} is employed for RDF2Vec. To ensure a fair comparison, all models are trained with default hyperparameters, except for the number of epochs, set to 100, and the embedding dimension, set to 200. 
Regarding implementation, the PyKEEN package\footnote{\url{https://pykeen.readthedocs.io/en/stable/}} is used to train TransE, TransR, DistMult, HolE, and BoxE, while pyRDF2Vec\footnote{\url{https://pyrdf2vec.readthedocs.io/en/latest/}} is employed for RDF2Vec. To ensure a fair comparison, all models are trained with default hyperparameters, except for the number of epochs, set to 100, and the embedding dimension, set to 200.

%%%%%%%%%%%%%%%%%%%%%%%% BIO-KGVEC2GO %%%%%%%%%%%%%%%%%%%%%%%%
\section{Bio-KGvec2go}

This paper presents a framework that collects new versions of biomedical ontologies, computes embeddings, and makes them publicly available through Bio-KGvec2go, \url{www.bio.kgvec2go.org}. The framework is designed to support an automated update mechanism that periodically downloads ontology releases from predefined URLs, computes checksums, and compares them with those of previously stored versions. If a change is detected, all embeddings are recomputed and made available.

\begin{figure}[h!]
    \vspace{0.3cm}
    \begin{subfigure}[b]{0.45\textwidth}
        \centering
        \includegraphics[width=0.95\textwidth]{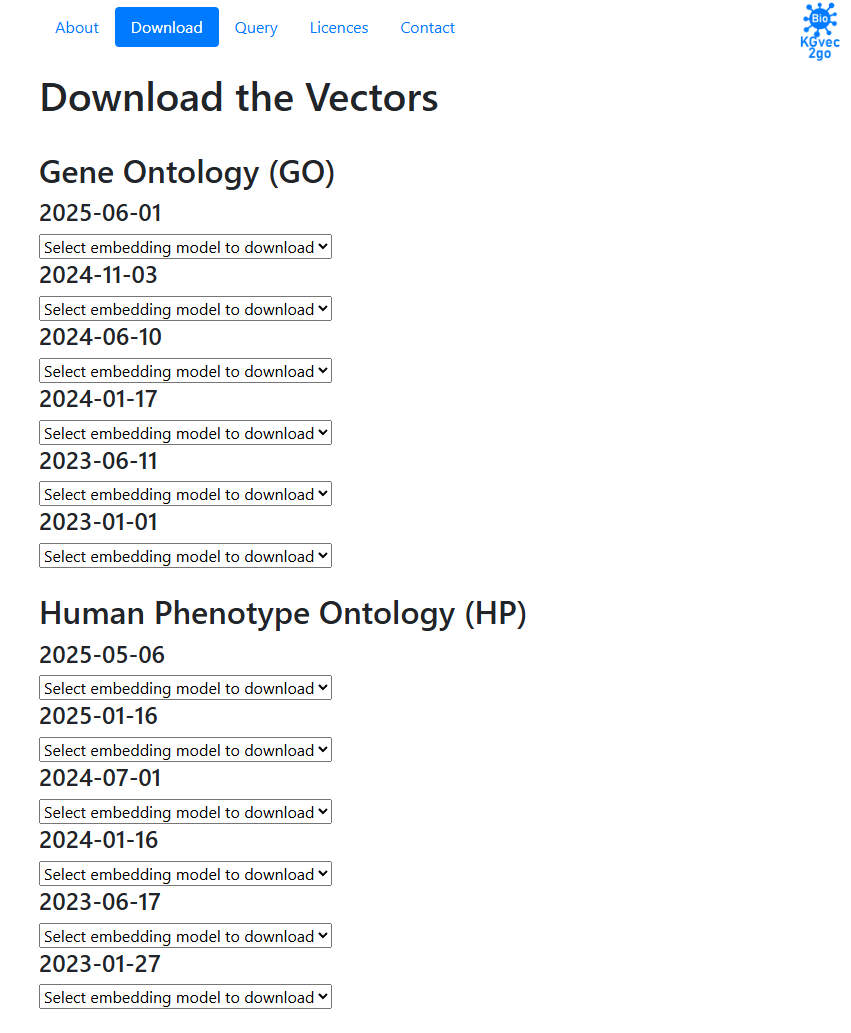}
        \caption{Download}
        \label{fig:download}
    \end{subfigure}
    \hfill
    \begin{subfigure}[b]{0.45\textwidth}
        \centering
        \includegraphics[width=0.95\textwidth]{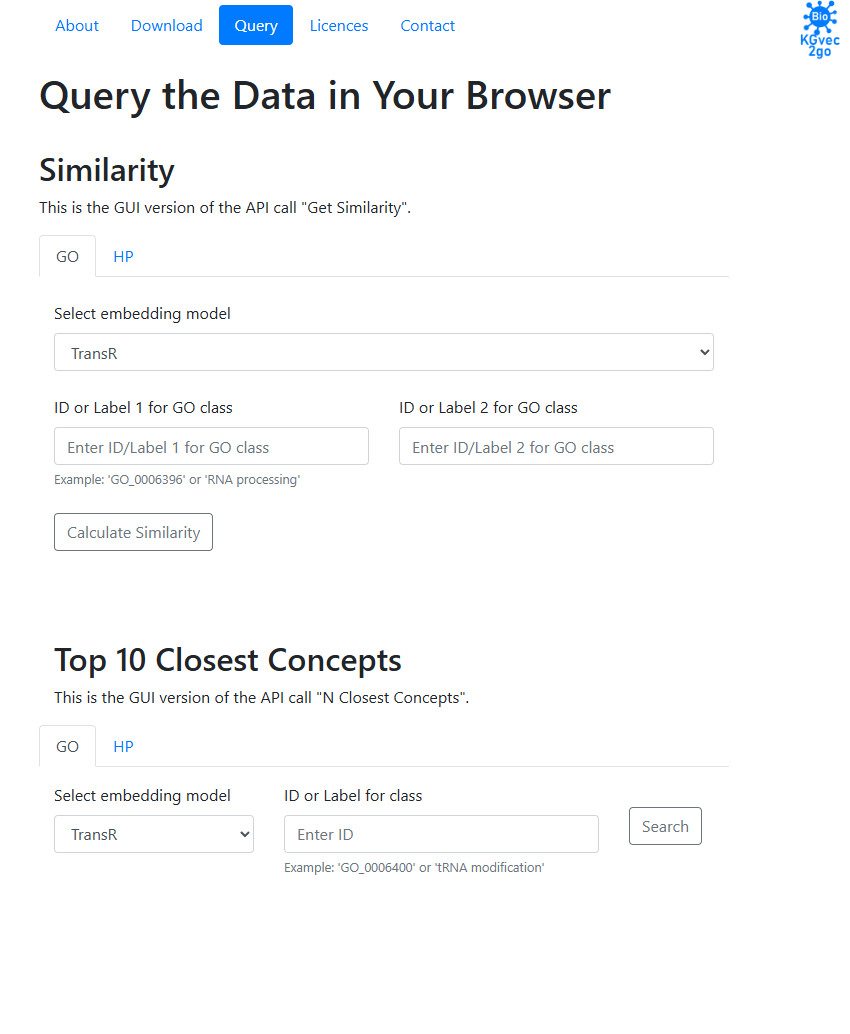}
        \caption{Similarity and Top Closest Concepts}
        \label{fig:query}
    \end{subfigure}
    \caption{Functionalities of Bio-KGvec2go.}
    \label{fig:functionalities}
\end{figure}

Regarding the platform Bio-KGvec2go, it is built upon the existing Web API, KGvec2go~\cite{portisch2020kgvec2go}, to provide access to up-to-date embedding models trained on biomedical ontologies. KGvec2go is implemented in Python using Flask and can be deployed with the Apache HTTP Server. The KGvec2go API, \url{www.kgvec2go.org}, offers a RESTful service designed to operate efficiently on Internet-connected devices with limited CPU and RAM (e.g., smartphones). Currently, Bio-KGvec2go offers three functionalities, as shown in Figure~\ref{fig:functionalities}: (i) downloading the embeddings for the various ontology versions, (ii) computing semantic similarity between two classes, and (iii) retrieving the top 10 most similar classes for any given ontology class. 

All the code is available on GitHub\footnote{\url{https://github.com/ritatsousa/biokgvec2go}}. Additionally, the trained KGE models are available on Zenodo\footnote{\url{https://zenodo.org/records/15865665}}, ensuring long-term preservation. The models are accompanied by metadata using the PROV standard~\cite{moreau2022provenance}, describing the input ontology, the KGE model used, and the corresponding hyperparameters.

\paragraph{Download}
Users can select an embedding model and download a corresponding JSON file containing the vector representations for each ontology class, encoded as 200-dimensional floating-point arrays. As of now, Bio-KGvec2go hosts embeddings for six distinct versions of each biomedical ontology, with the first version dating back to 2023 and subsequent versions released approximately every six months. Besides the downstream use of the embeddings, this functionality enables researchers to compare embeddings across different ontology snapshots, supporting studies on ontology evolution.
\vspace{-0.3cm}
\paragraph{Similarity}
Users can access the semantic similarity between two ontology classes by selecting an embedding model and providing either class identifiers or textual labels (with automatic normalization of case and whitespace). Bio-KGvec2go retrieves the corresponding vectors of the two classes from the most up-to-date version and computes the cosine similarity. The resulting score, ranging from -1 to 1, indicates the degree of similarity, where 1 denotes perfect similarity and -1 reflects complete dissimilarity. This functionality is particularly useful for ontology curation and annotation.
\vspace{-0.3cm}
\paragraph{Top Closest Concepts}
Users can select an embedding model and specify the target class by the identifier or label (with automatic normalization) to obtain the top 10 most semantically similar ontology classes. Bio-KGvec2go retrieves the most up-to-date version of embeddings and computes cosine similarities between the input vector and all other class vectors, returning a ranked list. The output is presented as a detailed table listing each related class by its identifier and label, the similarity score, and a direct URL for further exploration. This functionality is well-suited for semantic search and identification of candidates for enrichment analyses.

%%%%%%%%%%%%%%%%%%%%%%%% USE CASES %%%%%%%%%%%%%%%%%%%%%%%%
\section{Use Cases}

Bio-KGvec2go has been designed as a user-friendly platform, and it can support a broad spectrum of biomedical research. 
One key use case is in ontology-based ML approaches, where the embeddings can serve as input features for predictive models across diverse biomedical tasks. These approaches have become increasingly popular, as they allow the integration of structured biological knowledge.
For example, GO embeddings have been used to predict the function of uncharacterized proteins or to identify which proteins are likely to interact with each other. These interactions are crucial for many functions in biology and are highly relevant to disease states. Similarly, HP embeddings have been employed to uncover new associations between genes and diseases, improving the understanding of disease mechanisms. HP embeddings have also been used to improve disease diagnosis by comparing a patient's phenotypic profile to known disease profiles represented in the embedding space.
Since discovering protein interactions or gene-disease associations through laboratory experiments is expensive and time-consuming, ML-based approaches help to generate candidate pairs, narrowing the search space for lab validation and substantially reducing both the time and cost of experimental research.

% Ontology developement and semantic annotation
Another important application lies in ontology development, curation, and semantic annotation. Both GO and HP are manually curated by domain experts, many of whom have limited computational experience and therefore benefit from accessible tools. The \textit{similarity} and \textit{top closest concepts} functionalities provided by Bio-KGvec2go are particularly useful for these tasks. For instance, when annotating a gene with a specific function, researchers can use the tool to find the most semantically similar GO terms, ensuring more accurate and consistent annotations. Beyond annotation, this tool can help identify gaps or inconsistencies in the ontology itself. 

% In practical context in the lab (for protein development, drug design)
% Finally, Bio-KGvec2go can support laboratory research by guiding experimental design and hypothesis generation.
% In practical settings such as protein engineering or drug discovery, the tool can help identify candidate proteins and genes, narrowing the search space for experimental validation. This capability can substantially reduce both the time and cost of experimental research.

%%%%%%%%%%%%%%%%%%%%%%%% CONCLUSION %%%%%%%%%%%%%%%%%%%%%%%%
\section{Conclusion}

This paper presents Bio-KGvec2go, a FAIR resource designed to provide up-to-date pre-trained biomedical embeddings. 
Bio-KGvec2go extends the original KGvec2go API by incorporating multiple KGE models beyond RDF2Vec and by supporting several versions of the same biomedical KGs. By democratizing access to these embeddings, Bio-KGvec2go enables researchers to accelerate experimentation and improve performance across various tasks, including disease prediction, identification of gene-disease associations, and drug discovery. Moreover, by facilitating the reuse of pre-trained embeddings, it contributes to reducing the carbon footprint. 

As future work, we plan to expand Bio-KGvec2go to support additional biomedical KGs and embedding models. We also aim to improve the similarity and top closest concepts search functionalities by introducing features such as autocomplete for concept labels and tolerance to minor typos, ensuring that users can retrieve relevant concepts even if the input is not an exact match.

%%%%%%%%%%%%%%%%%%%%%%%% ADDITIONAL SECTIONS %%%%%%%%%%%%%%%%%%%%%%%%
\section*{Acknowledgments}
This work was funded by the Open Science Office of the University of Mannheim.

\section*{Declaration on Generative AI}

During the preparation of this work, the authors used ChatGPT and Grammarly for grammar checks, paraphrasing, and rewording. After using these tools, the authors reviewed and edited the content as needed and take full responsibility for the publication’s content.

%%%%%%%%%%%%%%%%%%%%%%%% BIBLIOGRAPHY %%%%%%%%%%%%%%%%%%%%%%%%
%% Define the bibliography file to be used
\bibliography{sample-ceur}

\end{document}